\documentclass[conference]{IEEEtran}
\IEEEoverridecommandlockouts

\usepackage{cite}                       
\usepackage{amsmath,amssymb,amsfonts}
\usepackage{algorithmic}
\usepackage{graphicx}
\usepackage{textcomp}

\usepackage[utf8]{inputenc}
\DeclareUnicodeCharacter{03BB}{$\lambda$}
\DeclareUnicodeCharacter{03C1}{$\rho$}
\DeclareUnicodeCharacter{03C3}{$\sigma$}
\DeclareUnicodeCharacter{2192}{$\rightarrow$}
\DeclareUnicodeCharacter{21D2}{$\Rightarrow$}
\DeclareUnicodeCharacter{2212}{$-$}
\DeclareUnicodeCharacter{00B1}{$\pm$}
\DeclareUnicodeCharacter{2248}{$\approx$}
\DeclareUnicodeCharacter{00D7}{$\times$}
\DeclareUnicodeCharacter{2264}{$\leq$}
\DeclareUnicodeCharacter{2265}{$\geq$}
\DeclareUnicodeCharacter{2713}{\checkmark}
\DeclareUnicodeCharacter{2014}{\textemdash{}}
\DeclareUnicodeCharacter{2013}{--}
\DeclareUnicodeCharacter{2020}{\dag}
\DeclareUnicodeCharacter{2022}{\textbullet}
\usepackage{xcolor}

\PassOptionsToPackage{disable}{todonotes}
\usepackage{papermacros}                

\def\BibTeX{{\rm B\kern-.05em{\sc i\kern-.025em b}\kern-.08em
    T\kern-.1667em\lower.7ex\hbox{E}\kern-.125emX}}

\usepackage[hidelinks]{hyperref}        
\hypersetup{pdfauthor={Naoto Nishida, Yoshio Ishiguro},pdftitle={Orthogonal Ensembles and Tested Explanations for Performer-Independent Body-Motion Emotion Recognition},pdfsubject={MMAC Challenge @ ACII 2026},pdfkeywords={affective computing, emotion recognition, body movement, skeleton motion, model ensemble, explainability, Laban Movement Analysis}}

\begin{document}

\title{Orthogonal Ensembles and Tested Explanations for
Performer-Independent Body-Motion Emotion Recognition}

\author{\IEEEauthorblockN{Naoto Nishida}
\IEEEauthorblockA{\textit{The University of Tokyo}\\
Tokyo, Japan\\
nawta@g.ecc.u-tokyo.ac.jp}
\thanks{This work was supported by JST BOOST Grant JPMJBS2418, JST
Moonshot R\&D Grant JPMJMS2012, JST CREST Grant JPMJCR17A3, the
commissioned research by NICT Japan Grant JPJ012368C02901, Tateisi
Science and Technology Foundation (C) and Telecommunications Advancement
Foundation.}
\thanks{Accepted for publication in the 2026 14th International Conference
on Affective Computing and Intelligent Interaction Workshops and Demos
(ACIIW). \copyright{} 2026 IEEE. Personal use of this material is permitted.
Permission from IEEE must be obtained for all other uses, in any current or
future media, including reprinting/republishing this material for advertising
or promotional purposes, creating new collective works, for resale or
redistribution to servers or lists, or reuse of any copyrighted component of
this work in other works.}
\and
\IEEEauthorblockN{Yoshio Ishiguro}
\IEEEauthorblockA{\textit{The University of Tokyo}\\
Tokyo, Japan\\
ishiy@acm.org}}

\maketitle

\begin{abstract}
We study body-only, 12-class acted-emotion classification from skeleton motion
under leave-performer-out (LPO) evaluation, a hard, underdetermined setting:
chance is $8.3\%$, and a protocol-matched reproduced STGCN++ baseline reaches
only $25.73 \pm 4.03\%$ Macro-F1. We show that reliable gains come not from a
new architecture but from combining eleven models with orthogonal error modes:
under 10-fold LPO cross-validation on the labeled training performers, an
equal-weight logit-mean ensemble reaches $36.80 \pm 4.00\%$ per-fold Macro-F1,
a protocol-matched $+11.07$~pp ($+43\%$ relative) over the same-split
reproduced baseline. Our central contribution is a tested explanation suite:
for a strong ensemble member, part-masking and counterfactual edits show
(rather than assert) that its decisions \emph{depend} on motion-grounded
body-region evidence, and this region saliency \emph{aligns} with rule-based
Laban Movement Analysis (LMA) attributes far more than with classical
kinematics: region-level saliency--LMA Spearman $\rho = +0.500$ versus
$+0.033$, roughly $15\times$, and the alignment holds for the submitted
11-way ensemble itself at $\rho = +0.517$; the audit is post hoc and needs
no retraining.
The same suite faithfully reports a negative: within-window temporal saliency
is diffuse rather than localized.
On the hidden challenge test set the submitted ensemble scored $37.23\%$
Macro-F1 and received the Best Performance Award of the MMAC Challenge
2026. Code is available at \url{https://github.com/nawta/diema-challenge}
and the presentation at \url{https://nawta.github.io/mmac2026/}.

\end{abstract}


\section{Introduction}
\label{sec:introduction}

The DIEM-A challenge~\cite{cheng2025diema} asks for 12-class \emph{acted} emotion recognition from
\emph{body motion alone}: no face, no audio, no scene, evaluated across
\emph{disjoint performers}. Together these make it underdetermined: chance is
$8.3\%$, and a strong skeleton baseline (STGCN++) reaches only $\sim$$25\%$
Macro-F1 under leave-performer-out (LPO) evaluation. Semantically related
emotions share kinematic signatures (e.g.\ jealousy/contempt, shame/guilt)
and each performer carries a movement idiolect, so cross-performer
within-class variance can rival within-performer cross-class variance.

Within this setting, intuitive single-model improvements largely do not
transfer: supervised-contrastive heads, VLM distillation, post-hoc
calibration, and other single-model upgrades failed or regressed under the protocol
(\secref{method}, \tbref{table2_negatives}); these negatives shaped a
deliberately conservative system.

What did transfer was diversity. Because four inductive-bias families
(graph, attention, hybrid/MLP, external pretraining) make \emph{different}
mistakes, averaging their raw logits cancels errors that any single family
repeats. This equal-weight \emph{logit-mean} ensemble, with skeleton
self-supervised pretraining, yields $36.80 \pm 4.00\%$ per-fold Macro-F1
under our study protocol, a protocol-matched $+11.07$~pp over the same
74-performer reproduced STGCN++ baseline, and the gain tracks the measured
orthogonality of member errors, not any one architecture
(\secref{performance_results}).

Our central methodological contribution: we \emph{test} attributions rather
than assert them. For a strong ensemble member, part-masking and
counterfactual edits show its decisions \emph{depend} on motion-grounded
body-region evidence, and this region saliency \emph{aligns} with rule-based
Laban Movement Analysis (LMA) attributes far more than with classical
kinematics, at saliency--LMA $\rho=+0.500$ vs $+0.033$, a post-hoc audit
needing no retraining; the same alignment holds for the submitted 11-way
ensemble itself. The suite returns six explicit verdicts, five positive
and one deliberately reported negative, plus a deterministic
$0/50$-audited narrator (\secref{explainability_results}).

\noindent\textbf{Contributions.}
\begin{itemize}[leftmargin=1.2em,itemsep=0.15em,topsep=0.2em]
  \item A \emph{tested} explanation suite for skeleton emotion models
        (part-masking faithfulness, perturbation stability, semantic LMA
        alignment, counterfactual edits, and a narration audit), returning
        six explicit verdicts: five positive, one reported negative.
        Part-masking shows a strong
        ensemble member's decisions \emph{depend} on motion-grounded
        body-region evidence, and that region saliency \emph{aligns} with
        rule-based LMA attributes (saliency--LMA $\rho=+0.500$ vs $+0.033$),
        a post-hoc audit needing no retraining
        (\secref{explainability_results}).
  \item The same suite faithfully reports a negative (diffuse temporal
        saliency) and a deterministic, audited motion$\rightarrow$rationale
        narrator; we contribute the \emph{method and audit}, not a dataset
        release (held pending consent/licen\-se review).
  \item A protocol-matched performance result: an orthogonal-error 11-way
        logit-mean ensemble, $+11.07$~pp / $+43\%$ over the same-split
        reproduced baseline, with the gain \emph{explained} by measured
        error-space orthogonality (\secref{performance_results}).
  \item A documented catalog of negative results
        (\tbref{table2_negatives}) that motivated the conservative design.
\end{itemize}

\begin{figure}[t]
\centering
\includegraphics[width=0.75\columnwidth]{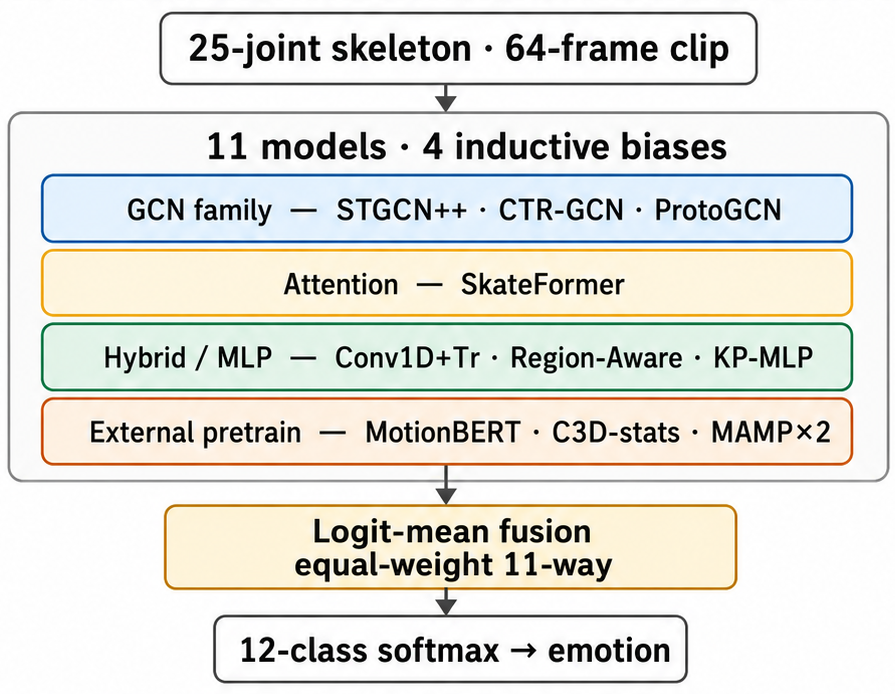}
\caption{The DIEM-A 12-emotion
body-motion challenge ($74$ train / $18$ test performer-disjoint performers;
BVH/FBX/C3D + text scenarios) and the proposed pipeline: eleven models
spanning four inductive biases (GCN, attention, hybrid/MLP, external
pretraining) fused by equal-weight logit-mean, reaching
$36.80 \pm 4.00\%$ per-fold Macro-F1 versus the same-split reproduced
STGCN++ baseline ($25.73 \pm 4.03\%$). Protocol: 10-fold leave-performer-out
on the 74-performer training split; per-fold mean $\pm$ SD throughout.}
\label{fg:system_overview}
\vspace{-2ex}
\end{figure}

\noindent\fgref{system_overview} sketches the task and pipeline; the
paper develops the two headlines above.

\section{Related Work}
\label{sec:related_work}

\noindent\textbf{Skeleton-based recognition and body emotion.}
Graph-convolutional and attention architectures dominate skeleton action
recognition (ST-GCN~\cite{yan2018stgcn}, CTR-GCN~\cite{chen2021ctrgcn},
STGCN++ and pose-heatmap variants~\cite{duan2022pyskl},
SkateFormer~\cite{do2024skateformer}); body-emotion datasets and
emotion-from-motion analyses extend the action setting to acted
affect~\cite{liu2022beat,fourati2014emilya,karg2013body,aristidou2015lma}.
These models set the single-network state of the art we build on but are
typically evaluated within-performer; we report the harder
leave-performer-out (LPO) regime.

\noindent\textbf{Self-supervised skeleton pretraining.}
Masked motion prediction~\cite{mao2023mamp} and contrastive skeleton
self-supervision~\cite{wu2023skeletonmae,thoker2021skeleton} transfer
representations across datasets, and unified motion encoders pretrained on
heterogeneous corpora~\cite{zhu2023motionbert} extend this transfer beyond
single-task supervision. We treat frozen external pretraining as one
orthogonal inductive-bias family within an ensemble, and report when such
transfer does \emph{not} help (\tbref{table2_negatives}).

\noindent\textbf{Explainability for motion models.}
Prior work explains motion and skeleton classifiers through perturbation and
occlusion~\cite{zeiler2014visualizing}, gradient-based
saliency~\cite{selvaraju2017gradcam}, counterfactual
edits~\cite{wachter2018counterfactual}, and semantic motion attributes
(e.g.\ Laban Movement Analysis~\cite{aristidou2015lma}); a parallel line
studies attribution \emph{faithfulness} and
\emph{stability}~\cite{hooker2019roar,adebayo2018sanity}. Most motion-domain
studies \emph{assert} that attributions are meaningful; few test whether
the cited evidence actually drives predictions, and fewer still under
performer shift. We contribute a \emph{validated audit protocol} rather
than a new attribution method: the probes above, screened for faithfulness
and stability and reported together with the negatives they return.

\section{Task, Data and Evaluation}
\label{sec:task_data_eval}

\noindent\textbf{Task and data.}
DIEM-A~\cite{cheng2025diema} is acted body-motion emotion recognition over 12
classes (anger, contempt, disgust, fear, joy, sadness, surprise, jealousy,
shame, guilt, gratitude, pride). Each clip is a 24-joint skeleton sequence
(BVH/FBX/C3D) with a text scenario; models consume a 25-node tensor (the 24
joints plus a virtual root node carrying global position), the resolution at
which our part-masking and saliency analyses are reported. Performers are
split $74$ train / $18$ test and are \emph{disjoint} across the
split~\cite{mmac2026benchmark}. The training split contains $7{,}992$ clips
from $40$ Japanese and $34$ Taiwanese performers ($666$ per emotion, balanced
across classes) and the test split $1{,}944$ clips from $18$ disjoint
performers ($9$ JP, $9$ TW); recordings are at $120$~Hz with median sequence
length $\approx 845$~frames, of which we feed every model a fixed $64$-frame
window ($\approx 7.5\%$ of the median, $\approx 0.53$~s) for protocol parity
with the official baseline.

\noindent\textbf{Capture provenance.}
The corpus was recorded with two motion-capture systems: the five earliest
Japanese performers (JP\_01--JP\_05) were captured with a 41-marker OptiTrack
rig, and all remaining recordings use the production 57-marker Vicon
system~\cite{cheng2025diema}. None of the five OptiTrack performers is in the 74-performer training
split, so capture hardware is uniform within our cross-validation; the
residual capture-\emph{era} stratum is checked in
\secref{performance_results} by excluding the seven earliest-captured
Japanese training performers (JP\_06--JP\_12).

\noindent\textbf{Evaluation.}
We evaluate with 10-fold leave-performer-out (LPO) cross-validation on the
74-performer training split and report Macro-F1 (primary) and Accuracy. We fix
\emph{one} reporting convention and use it everywhere: scores combine member
predictions by \emph{logit-mean} (mean of raw logits, the canonical
convention) and are reported as the \emph{per-fold mean $\pm$ SD across the 10
LPO folds}; pooled out-of-fold (OOF) values appear only as a clearly labelled
descriptive secondary. This convention is restated in every table and figure
caption (\tbref{table1_main_results}).\footnote{Per-fold reporting matches
the SD convention used by the official and reproduced STGCN++ baselines and
supplies the fold-level error bars and bootstrap intervals used throughout
($B{=}1000$, seed~42);
probability-domain aggregation is an internal ranking device only.}

\noindent\textbf{Protocol distinction (stated here, not deferred).}
The official STGCN++ baseline~\cite{mmac2026benchmark} is a 92-performer
full-LPO result ($25.21 \pm 4.49\%$ Macro-F1). Our development numbers use the 74-performer
training split (test labels are withheld by the challenge). \emph{Within the
same 74-performer protocol our reproduced STGCN++ baseline is
$25.73 \pm 4.03\%$ (within one SD of the official $25.21 \pm 4.49\%$), so
we treat the official number as an external anchor and report all
improvements against the protocol-matched reproduction}. Consequently the headline
improvement is the protocol-matched $+11.07$~pp ($36.80-25.73$, same split,
same logit-mean $\times$ per-fold convention), never a comparison to the
92-performer official figure. On the hidden test set the submitted
ensemble scored $37.23\%$ Macro-F1 and $37.50\%$ accuracy
(organizers' final leaderboard~\cite{mmac2026benchmark}; Best Performance
Award); this single score carries no fold-level spread, so the
cross-validated $36.80 \pm 4.00\%$ remains the headline and the test score
corroborates it.

\section{Method}
\label{sec:method}

\subsection{Model pool: four inductive-bias families}
\label{subsec:model_pool}
The pool has eleven members spanning four families with deliberately
\emph{different} inductive biases, so their errors are unlikely to coincide:
\one graph-convolutional (STGCN++~\cite{duan2022pyskl},
CTR-GCN~\cite{chen2021ctrgcn}, ProtoGCN~\cite{liu2025protogcn}, a
Region-Aware ConvTr); \two attention (SkateFormer~\cite{do2024skateformer});
\three hybrid/MLP (Conv1D+Transformer, Keypoint-Pool-MLP); \four frozen
external pretraining (MotionBERT-Lite~\cite{zhu2023motionbert},
C3D-marker-stats, MAMP NTU60-xsub and NTU120-xset~\cite{mao2023mamp}).
ProtoGCN is a CTR-GCN backbone with a learned per-body-part
prototype-matching head; Region-Aware ConvTr and the hybrid/MLP family are
part-aware variants of the same Conv1D+Transformer template. The four
families encode complementary priors: kinematic-chain \emph{locality},
\emph{long-range position-insensitive co-occurrence}, \emph{pooled
spatiotemporal statistics}, and \emph{transferred representations} from
generic-motion corpora. The explained model in
\secref{explainability_results} (Region-Aware, strongest single at
$30.05 \pm 3.48\%$) is itself a member of the submitted ensemble.

\subsection{Fusion: equal-weight logit-mean}
\label{subsec:fusion}
Given per-member logits $z_m \in \mathbb{R}^{12}$ for $M$ members, the
prediction is the argmax of the equal-weight mean of \emph{raw logits}:
\begin{equation}
\hat{y} \;=\; \arg\max_{c}\; \frac{1}{M}\sum_{m=1}^{M} z_m^{(c)} .
\label{eq:logitmean}
\end{equation}
Averaging raw logits before normalisation (a log-domain geometric mean of
class scores) keeps confidently-wrong members from dominating. Empirically
logit-mean fusion is $+1.15$~pp Macro-F1 over the probability-averaging
variant at no cost; we therefore fix logit-mean as the canonical convention
(\secref{performance_results}). 

\subsection{Pretraining and protocol discipline}
\label{subsec:protocol}
One family uses skeleton self-supervised pretraining (masked-motion, MAMP
style) transferred frozen: two MAMP backbones pretrained on NTU60-xsub and
NTU120-xset respectively, joined by a frozen MotionBERT-Lite encoder and a
C3D-marker-statistics encoder. Training and selection follow a strict protocol:
10-fold LPO, fixed seed, fold-wise out-of-fold (OOF) prediction, model
selection \emph{within} the training split only, and no access to test labels
(withheld by design); ensemble weights are equal (no fitted weighting).
In-domain members share the data split, batch size $128$, clip length $64$
and seed~$42$, but use model-tuned optimisers: STGCN++, CTR-GCN and
ProtoGCN train with SGD at lr~$0.2$ for $65$~epochs without warmup;
SkateFormer with AdamW at lr~$5{\times}10^{-4}$ for $65$~epochs with a
$5$-epoch cosine warmup; Region-Aware ConvTr, Conv1D+Transformer and
Keypoint-Pool-MLP with AdamW at lr~$\in\{1, 4\}\!\times\!10^{-3}$ for
$80$~epochs with the same warmup. The four frozen externals contribute
features and add $<\!0.01$~M trainable parameters per branch (a linear
classification head only). Each fold writes per-clip raw-logit out-of-fold
(OOF) arrays in the canonical $(n,12)$ shape that feeds
Eq.~\ref{eq:logitmean}.

\noindent\textbf{Computational cost and reproducibility.}
The system is deliberately cheap for an eleven-member ensemble. Seven
members are trained in-domain; the four external branches stay frozen with
a linear head each, so the total trainable footprint is $12.98$\,M
parameters (\tbref{table1_main_results}), roughly the seven-model sum.
Inference is eleven forward passes over a $64$-frame clip followed by the
parameter-free logit mean of Eq.~\ref{eq:logitmean}; there are no fitted
fusion weights, and a single GPU suffices for training any member and for
inference. The explanation suite runs post
hoc, at zero accuracy cost and with no retraining. Every number, table,
and figure in this paper is produced by a single deterministic
regeneration script over the stored per-member OOF arrays.

\subsection{Negatives as design constraints}
\label{subsec:negatives}
The conservative design above is what \emph{survived} a broad search; the
approaches that did not are reported as constraints, not omitted
(\tbref{table2_negatives}): supervised-contrastive, mixture-of-experts,
scenario-text alignment and VLM zero-shot/distillation, multi-crop window
inference, and post-hoc calibration each failed or regressed; C3D late
fusion ($-0.59$~pp) was associated with a $69.8\%$ country-leakage signal.
Effect sizes and CIs live in the table.

\begin{table*}[t]
\centering
\caption{Approaches that did not improve the system, each paired with its methodological lesson (non-causal wording: associated with / regresses / does not improve). $\Delta$F1 is the paired change versus the matched baseline. The full catalogue is in the supplementary material.}
\label{tab:table2_negatives}
\footnotesize
\begin{tabular}{l p{2.6cm} p{2.7cm} p{4.9cm}}
\toprule
Approach & Configuration & Outcome ($\Delta$F1) & Methodological lesson \\
\midrule
Supervised contrastive head & pairwise SupCon, $\lambda$=0.05, batch 128 & -11.53 pp, significant & 12 classes at batch 128 give too few positives per class, suggesting a per-class center loss may fit better at this scale \\
Mixture-of-experts fusion & K=4, hand-crafted gradient-free routing & collapses to 6.55\% F1 & a shared head with random init and 21-D routing is associated with backbone collapse, suggesting a feature adapter, learned routing, and zero-init are needed \\
Scenario-text alignment & global InfoNCE, 3-fold & -1.42 pp & the projection collapses to performer identity (cross-group retrieval 10.6\%) \\
Part-rationale alignment & cosine text alignment, 3-fold & -0.64 pp, CI [-1.30, -0.09] & the text bottleneck is associated with an F1 ceiling of 9--12\%; shortcut-breaking works but yields no F1 lift \\
VLM zero-shot distillation & 12-class, a recent VLM, preliminary & failed in preliminary zero-shot ranking & the true class ranked last (12/12); a vision-language model did not infer emotion reliably from faceless, scene-free skeleton renderings \\
Contact-marker late fusion & C3D 22-D contact stats, 3-fold & -0.59 pp & the contact features carry a 69.8\% country signal, so some performers go out of distribution under LPO, suggesting a domain-invariance objective is needed \\
Multi-crop window inference & 24-setting window sweep & -12 to -17 pp & the model is trained on full-span subsamples, so compact windows are a distribution shift \\
Post-hoc calibration & nested-LPO, 7-way ensemble & 0 to -0.14 pp, not significant & equal-weight argmax is already near-optimal; bias-style calibration did not exceed it in this search \\
\bottomrule
\end{tabular}
\end{table*}

\section{Performance Results}
\label{sec:performance_results}

\noindent\textbf{Main result.}
\tbref{table1_main_results} reports all systems under the single canonical
convention defined in \secref{task_data_eval}: logit-mean fusion, per-fold
mean $\pm$ SD, 10-fold LPO on the 74-performer split. The submitted 11-way
ensemble reaches $36.80 \pm 4.00\%$ per-fold Macro-F1, a protocol-matched
$+11.07$~pp ($+43\%$ relative) over the same-split reproduced STGCN++
baseline of $25.73 \pm 4.03\%$; the pooled-OOF value $36.94\%$ appears in
\tbref{table1_main_results} as a descriptive secondary. The bootstrap
$95\%$ CI $[35.90, 37.94]$ clears the 7-way ensemble CI, and the
paired-fold improvement excludes zero in $10/10$ folds with a bootstrap
$95\%$~CI of $[9.86, 12.33]$~pp. Resampling whole performers rather than
folds widens this interval only to $[9.14, 12.17]$~pp, so the gain is
neither within ensemble noise nor an artifact of within-performer clip
correlation.



\begin{figure}[!t]
\centering
\includegraphics[width=\columnwidth]{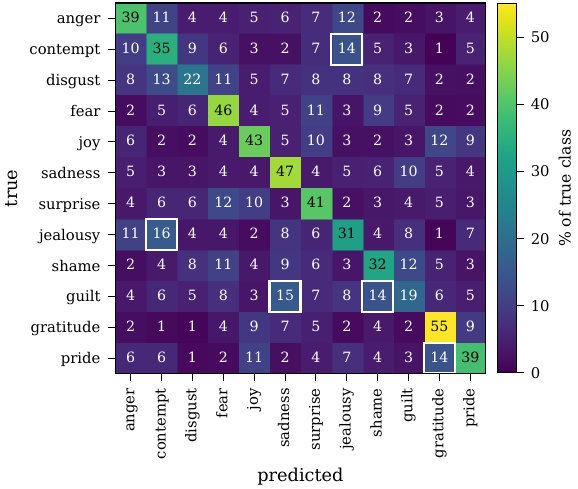}
\caption{$12\times12$ row-normalised pooled-OOF confusion of the 11-way
logit-mean ensemble ($n=7{,}992$ clips, 74-performer split; rows = true
class, rows sum to $100\%$); the five strongest off-diagonal confusions
are boxed. The per-class-by-country companion panel appears in the
supplementary material.}
\label{fg:confusion_jptw}
\end{figure}

\begin{figure}[t]
\centering
\includegraphics[width=0.9\columnwidth]{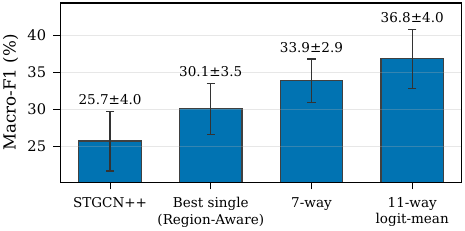}
\caption{Diversity across the four
families builds the gain: stage-by-stage Macro-F1 over the 11-model lift
path (10-fold LPO mean $\pm$ SD), baseline $\rightarrow$ best single
(Region-Aware) $\rightarrow$ 7-way $\rightarrow$ 11-way logit-mean.
Per-class progression and paired-bootstrap CIs are in the supplementary
material.}
\label{fg:lift_path}
\vspace{-2ex}
\end{figure}

\noindent\textbf{Lift path and ablations.}
\fgref{lift_path} traces the lift, under the one convention, as four stages:
$25.7 \rightarrow 30.1 \rightarrow 33.9 \rightarrow 36.8\%$ (baseline
$\rightarrow$ best single $\rightarrow$ 7-way $\rightarrow$ final 11-way
logit-mean). The intermediate $+0.7$~pp from frozen external pretraining
(MotionBERT-Lite, C3D-marker-stats) and $+2.2$~pp from two MAMP variants
are absorbed into the final stage, justified by member-level orthogonality
(quantified below). Two ablations matter. First, ensembling beats the best single model,
$30.05 \pm 3.48\%$, by $+6.8$~pp. Second, logit-mean beats
probability-averaging by a free $+1.15$~pp; the absolute value of the
probability-averaging convention serves only for internal ranking and is
not reported. This $+1.15$~pp is not a logit-scale artifact: z-scoring
each member to unit variance neutralises any single high-norm member yet
retains $+0.99$~pp, and rank-only Borda fusion retains $+0.69$~pp. All 12
classes improve over the baseline, the weakest (sadness) by $+6.1$~pp
(progression figure in the supplementary material).

\noindent\textbf{Why diversity helps: error-space orthogonality.}
The gain comes from disagreement, not from added capacity: the eleven
members rarely fail on the same clip. Their
pairwise per-sample error correlations are low, all off-diagonal values
inside $[0.15, 0.43]$ and lowest for the frozen external-pretraining
column, so averaging their logits cancels errors that a single family
would share (correlation matrix in the supplementary material). The most-redundant
pair, Conv1D+Tr\,$\leftrightarrow$\,Keypoint-Pool-MLP, and the elevated
MAMP-NTU60\,$\leftrightarrow$\,MAMP-NTU120 pair are consistent with the
small or negative return of further homogeneous additions
(\tbref{table2_negatives}). A leave-one-out (LOO)
check on the pooled-OOF logit-mean base of $36.94\%$ confirms no member
dominates the lift and no member is redundant: all eleven per-member deltas are strictly
negative, from $-0.84$~pp for MAMP-NTU60 down to $-0.09$~pp for CTR-GCN,
and all within the $7\rightarrow11$-way lift margin; the full per-member
table (with per-stratum versions) appears in the supplementary material. The
smallest losses come from graph-convolutional members whose errors a
same-family sibling largely covers (CTR-GCN $-0.09$, ProtoGCN
$-0.15$~pp), yet the moderately correlated MAMP pair still includes the
single largest contribution, $-0.84$~pp, with $-0.61$~pp for its sibling;
moderate error correlation therefore does not imply redundancy. Removing
the whole frozen external-pretraining block costs $-2.91$~pp, so
the externals jointly supply essentially the whole $7\rightarrow11$-way
lift, spread across branches rather than carried by one.

\noindent\textbf{Error structure and strata.}
\fgref{confusion_jptw} shows where the residual errors go: between
\emph{semantically confusable} emotions; the five strongest
off-diagonal confusions, boxed, include
jealousy\,$\rightarrow$\,contempt and guilt\,$\rightarrow$\,sadness/shame.
This pattern matches the model's low maximum confidence
($\sim$$62\%$) and the disagreement structure the ensemble exploits. Per-class F1 by performer
country is charted in the supplementary material. The JP--TW macro gap ($35.3$ vs
$38.9\%$) is a \emph{dataset stratum, not a cultural finding}: it is
entangled with capture era and performer idiolect. The five
OptiTrack-captured performers are absent from the training split
(\secref{task_data_eval}), so capture hardware cannot explain the
cross-validation gap; as a capture-era check, excluding the seven
earliest-captured Japanese training performers shifts pooled-OOF Macro-F1
by only $-0.44$~pp ($36.50$ vs $36.94\%$), so neither the headline nor the
stratum hinges on the earliest recordings.

\begin{table*}[t]
\centering
\caption{Main results on DIEM-A (10-fold LPO, 74-performer train split). Final row = submitted model. Macro-F1 / Accuracy = 10-fold LPO per-fold mean $\pm$ SD (the official convention); fusion = logit-mean (mean of raw logits); 95\% CI = sample-level paired bootstrap, 1000 iter, seed 42, on pooled OOF. Pooled-OOF F1 is a descriptive secondary: 7-way 34.03\%, 11-way logit-mean 36.94\%.}
\label{tab:table1_main_results}
\footnotesize
\begin{tabular}{l r c c c}
\toprule
System & Trainable params (M)$^{\ddagger}$ & Macro-F1 (mean $\pm$ SD) & Macro-F1 95\% CI & Accuracy (mean $\pm$ SD) \\
\midrule
STGCN++ official baseline & 1.40$\star$ & 25.21 $\pm$ 4.49$\star$ & n/a$\star$ & 27.11 $\pm$ 3.67$\star$ \\
STGCN++ reproduced & 1.41 & 25.73 $\pm$ 4.03 & [25.37, 27.23] & 27.54 $\pm$ 3.92 \\
Best single (Region-Aware) & 1.03 & 30.05 $\pm$ 3.48 & [29.32, 31.15] & 30.87 $\pm$ 3.43 \\
7-way ensemble & 12.98 & 33.86 $\pm$ 2.92 & [33.01, 35.04] & 34.68 $\pm$ 3.00 \\
\textbf{11-way logit-mean (submitted)} & \textbf{12.98} & \textbf{36.80 $\pm$ 4.00} & \textbf{[35.90, 37.94]} & \textbf{37.40 $\pm$ 4.06} \\
\bottomrule
\end{tabular}
\\[2pt] \parbox{\linewidth}{\raggedright\scriptsize $\star$ Challenge-reported (92-performer full LPO, official 25.2 \% $\pm$ 4.5 \%, no bootstrap CI), NOT re-evaluated with our script; our reproduction (25.73 $\pm$ 4.03) is within SD of the official number, so the official baseline serves only as an external anchor and all improvement numbers are same-split comparisons. \quad $^{\ddagger}$ The table reports trainable parameters; frozen external branches add fewer than 0.01 M trainable parameters each.}
\end{table*}

\section{Explainability Results}
\label{sec:explainability_results}

We do not just visualize what the model attends to; we \emph{test} it. For
every claim that the model uses a body region, we mask it, perturb it, or
edit its motion and check that the prediction moves. The suite runs on the
Region-Aware member of the submitted ensemble and its key tests re-run on
the ensemble itself; \tbref{table3_explainability} collects every value with
its CI or null and a plain-language reading. Six verdicts, five
positive and one deliberately negative:
\textbf{faithful} (masking the parts ranked important degrades Macro-F1,
\S\ref{subsec:faithfulness}); \textbf{stable} (the ranking survives input
noise, \S\ref{subsec:stability}); \textbf{semantically grounded} (region
saliency matches the Laban vocabulary, for the member and the submitted
ensemble alike, \S\ref{subsec:lma}, \S\ref{subsec:ensemble_faithful});
\textbf{behaviourally corroborated} (motion edits shift predictions in the
same part ranking, \S\ref{subsec:counterfactual}); \textbf{grounded
narration} (no unsupported claims in the $50$ audited cards,
\S\ref{subsec:cards}); and \textbf{one reported negative} (within-window
temporal saliency is diffuse, \S\ref{subsec:negatives_explain}).
The suite is post hoc and adds no accuracy cost; its load-bearing LMA
headline rests on the preceding faithfulness and stability verdicts.

\begin{table*}[t]
\centering
\caption{Explainability metrics, with a plain reading of each row in the last column. Six rows are positive evidence (faithfulness, stability, counterfactual, narrator, submitted ensemble, LMA); temporal saliency is a deliberately reported negative. All values are recomputed on the same 10 LPO folds and out-of-fold predictions unless otherwise stated (10/10 folds for part-masking/stability). Bold = headline contrast.}
\label{tab:table3_explainability}
\footnotesize
\begin{tabular}{l p{2.5cm} p{2.9cm} p{2.1cm} p{3.9cm}}
\toprule
Component & Metric & Value & 95\% CI / null & What it shows \\
\midrule
Part-masking & Faithfulness AUC gap (important-reverse), zero-mask & +0.124 $\pm$ 0.031 & 10-fold SD; positive in all folds & Masking important parts lowers F1: attributions are faithful \\
Temporal saliency & AUC gap (salient-reverse), negative result & +0.002 ($\approx$0) & entropy 98.7\% of log T ($\approx$uniform)$^{\dagger}$ & Frame importance is diffuse (a reported negative) \\
Stability & Spearman $\rho$ vs $\sigma$=0 ref at $\sigma$=0.02 & +0.983 $\pm$ 0.039 & 10-fold SD; top part never flips & Attribution ranking is stable under input noise \\
Counterfactual & Most-disruptive edit $\Delta$p(true): head+freeze & -0.0164 & mean over 864 val $\times$ 10 folds & Perturbing motion changes predictions: motion-dependent \\
Narrator grounding & Unsupported claims in audited cards & 0/50 cards & deterministic audit & No unsupported claims in the 50 audited cards \\
F1 cost & $\Delta$F1: baseline $\rightarrow$ with post-hoc saliency/LMA export & 0.0 pp & post-hoc; no retraining & Explanations add no accuracy cost \\
Submitted ensemble & Part-masking + LMA under a shuffle (permutation-importance) mask & part gap +7.85 $\pm$ 1.76 pp; $\rho$ = +0.517 & part gap 10-fold SD; $\rho$ 95\% CI [+0.333, +0.700]; permutation $p = 0.001$ & Part-masking faithfulness and LMA alignment also hold for the submitted 11-way logit-mean ensemble (baseline Macro-F1 34.65 $\pm$ 4.29\%), not just one model \\
\textbf{LMA correlation} & \textbf{Region-level saliency--LMA Spearman $\rho$} & \textbf{$\rho$ = +0.500; kinematics control $\rho$ = +0.033} & \textbf{permutation $p<0.001$} & \textbf{Semantic alignment with the Laban vocabulary} \\
\bottomrule
\end{tabular}
\\[2pt] \parbox{\linewidth}{\raggedright\scriptsize $^{\dagger}$ Per-sample entropy; the per-emotion class-mean is 99.5\% of log T --- two aggregations of the same near-uniform distribution.}
\end{table*}

\subsection{Faithfulness: masking the cited regions lowers accuracy}
\label{subsec:faithfulness}
\emph{Positive.} Masking the parts an attribution calls important degrades Macro-F1 more than
masking the same number of ``unimportant'' parts: the important$-$reverse AUC
gap is $+0.124 \pm 0.031$ over 6 body parts ($+0.199$ at 25-joint
resolution), positive in every fold. Nothing downstream would matter if
attributions were decorative; this shows they are not.

\subsection{Stability: the ranking survives input noise}
\label{subsec:stability}
\emph{Positive.} Under input noise the attribution ranking is preserved (Spearman
$\rho=+0.983 \pm 0.039$ at $\sigma=0.02$; the top part never flips),
so \S\ref{subsec:faithfulness} is a property, not an artifact.

\subsection{What the model reads: body motion in Laban terms}
\label{subsec:lma}
\emph{Positive: the load-bearing result.} The model reads body motion the
way a movement analyst would name it. Region-level saliency aligns with
rule-based Laban Movement Analysis (LMA) attributes far more than with
classical kinematics (Spearman $\rho=+0.500$ vs $+0.033$, roughly $15\times$
stronger); for sadness, the salient regions track a bowed, sunken posture.
The $+0.500$ is computed over the $48$ emotion$\times$region
pairs. These pairs share a repeated $12\times4$ structure and are not $48$
independent observations, so the interval estimate treats emotions as the
resampling unit: the $95\%$ CI of $[+0.150, +0.733]$ is an emotion-block
bootstrap over the $12$ emotion clusters, and the permutation null itself
respects the structure, shuffling the four region scores \emph{within}
each emotion ($p<0.001$, null mean $\rho\approx 0$). A cluster-level check
agrees: the per-emotion alignment is positive for $10$ of the $12$
emotions, with median per-emotion $\rho=+0.70$ (exact sign test over
emotions, two-sided $p=0.039$).
The rule-based export needs no model retraining, so the Macro-F1 cost is
$0$~pp by construction (the bold row of \tbref{table3_explainability});
per-emotion LMA z-scores and the counterfactual deltas appear in the
supplementary material. The order-of-magnitude gap over the classical-kinematics comparator
anchors the body evidence in an established movement vocabulary rather
than leaving it merely self-consistent.

\noindent\textbf{How the per-emotion signature is built.}
Each clip yields a rule-based LMA attribute vector $a\in\mathbb{R}^{32}$ over
the four Laban axes (Body, Effort, Shape, Space; the full 32-attribute
schema is tabulated in the supplementary material).
Each attribute $k$ is standardised against the whole training pool and
averaged within an emotion, giving a per-emotion \emph{signature}: for
emotion $c$,
\begin{equation}
z_{c,k}=\frac{\bar a_{c,k}-\mu_k}{\sigma_k},\qquad
\bar a_{c,k}=\frac{1}{|C_c|}\sum_{i\in C_c} a_{i,k},
\label{eq:lma_z}
\end{equation}
where $\mu_k,\sigma_k$ are the global mean and SD of attribute $k$ over all
clips and $C_c$ is the set of clips of emotion $c$, so $z_{c,k}$ is how many
SD attribute $k$ departs from the corpus norm for that emotion. Grouping the
32 attributes into four body regions (head, arms, legs, torso; index sets
$R_r$) gives a region-importance vector
$\ell_{c,r}=\tfrac{1}{|R_r|}\sum_{k\in R_r}|z_{c,k}|$.
The headline $\rho=+0.500$ is computed by flattening the 12 emotion $\times$
4 region pairs and measuring the Spearman rank correlation between the LMA
region score and the model's region saliency; the classical-kinematics
control (per-joint speed, acceleration, range of motion, and energy) under
the identical pipeline reaches only $+0.033$. The signature is human-readable: sadness loads on
\texttt{head\_bow}$+$, \texttt{head\_height}$-$, and \texttt{trunk\_lean}$+$,
a bowed, sunken posture that the model's salient regions track. We treat this LMA correlation as semantic alignment,
not causal evidence by itself; dependence is tested separately through
part-masking and counterfactual perturbations
(\S\ref{subsec:faithfulness}, \S\ref{subsec:counterfactual}).

\subsection{The finding holds for the submitted ensemble}
\label{subsec:ensemble_faithful}
\emph{Positive, for the submitted system.} Because these probes target a
single member, we re-run the part-masking
and LMA-alignment tests on the submitted 11-way logit-mean ensemble
itself, using permutation-importance shuffling (zero-masking collapses
members that lack per-part gates). The important$-$reverse faithfulness gap stays
positive in $10/10$ folds ($+8.5$~pp partial-AUC at single-part
resolution), and region-level part importance aligns with the rule-based
LMA attributes at Spearman $\rho=+0.517$, with an emotion-block bootstrap
$95\%$ CI of $[+0.333, +0.700]$ and permutation $p=0.001$,
\emph{exceeding} the single-member $+0.500$; at the cluster level the
per-emotion alignment is positive for $11$ of the $12$ emotions and
negative for none (exact sign test, two-sided $p=0.001$). The
gap is smaller than the member's $+12.4$~pp, since six of the seven
skeleton members lack per-part gates, but it is consistently positive and
above null: the body-evidence finding describes the system we submit, not
one component.

\subsection{Counterfactual edits move predictions as predicted}
\label{subsec:counterfactual}
\emph{Positive, as observational corroboration.} Pose-preserving,
motion-perturbing edits shift logits in the \emph{same} part
ranking as masking: a head-freeze edit is the most disruptive, shifting the
true-class probability by $\overline{\Delta p_{\text{true}}}=-0.0164$ on
average over $864$ validation clips $\times$ $10$ folds and flipping
$41\%$ of predictions, and an arm-amplify edit flips $23.1\%$.
This is convergent \emph{observational} evidence (not causal identification)
that the model relies on motion dynamics. The per-part saliency-vs-disruption
agreement is positive but modest, at $\rho=+0.49$ with $n=6$ and $p=0.33$,
and we treat it as suggestive corroboration; the load-bearing
cross-method result remains the $\rho=+0.500$ vs $+0.033$ contrast with
its real $n$.

\subsection{What the explanations correctly refuse to claim}
\label{subsec:negatives_explain}
\emph{Negative: reported, by design.} Temporal saliency is
near-uniform for all 12 emotions (per-sample entropy $\approx 98.7\%$ of
$\log T$; the class-mean is $99.5\%$, two aggregations of the same
near-uniform distribution, reconciled in \tbref{table3_explainability}),
with an important$-$reverse AUC gap of only $\approx +0.002$: the model's
strongest evidence is \emph{spatial}: within-window frame
importance is diffuse (no single frame dominates). We report this negative rather than hide it; it does
not claim that temporal order carries no signal (saliency grids are in
the supplementary material). Reverse/random masking separates cleanly from important-first. A
faithfulness suite that only ever returned positive results would be
unfalsifiable; reporting these negatives is what makes it a test.
Per-row CIs/markers live in \tbref{table3_explainability} and
\tbref{table2_negatives}.

\noindent\textbf{One fused thesis: why diversity helps, explained.}
Inter-performer F1 varies $\sim$$4\times$ and decomposes into roughly half
intrinsic difficulty and half architecture-specific error, the same
disagreement structure the ensemble exploits (\secref{performance_results}):
one mechanism behind both headlines.

\subsection{Scoped qualitative cards (decoupled from the audit)}
\label{subsec:cards}
Narration is only a scoped interface to the audited saliency and LMA
channels; no quantitative claim relies on it. A label-field audit of the
cards is in the supplementary material.

\section{Discussion and Limitations}
\label{sec:discussion}

The following limitations are central to interpreting the benchmark; each
carries the reason the load-bearing claim still stands. The system is a
research probe under a fixed protocol, not a deployable affect recognizer.

\noindent\textbf{Absolute accuracy.} Macro-F1 is $\sim$$37\%$, but the task
is 12-way body-only LPO acted-emotion recognition (chance $8.3\%$; official
baseline only $25\%$); the contribution is a protocol-matched $+43\%$
relative gain with all 12 classes improving (weakest $+6.1$~pp), not
saturation.

\noindent\textbf{Protocol gap.} Our $36.80 \pm 4.00\%$ is a 74-performer cross-validation
result; under the same protocol our reproduced baseline is
$25.73 \pm 4.03\%$, within one SD of the official figure
(\secref{task_data_eval}), so the lift is protocol-matched; the
hidden-test $37.23\%$ Macro-F1 lies within the cross-validated spread.

\noindent\textbf{Temporal scope.} Every model sees a $64$-frame
window for parity with the official baseline (\secref{task_data_eval});
longer and multi-window inference regressed
(\tbref{table2_negatives}), so our claims concern this official
short-window LPO setting and the within-window evidence it exposes, not
full-sequence affect understanding.

\noindent\textbf{Variance and confidence.} Per-fold SD is $\approx 4$~pp,
but the bootstrap $95\%$ CIs (\tbref{table1_main_results}) show the
ensemble interval clearing the 7-way interval, so the gain is not
noise; the low maximum confidence ($\sim$$62\%$) is the
sensible response to confusable emotions, and what the ensemble exploits.

\noindent\textbf{Inter-performer range.}
Inter-performer F1 varies $\sim$$4\times$; this bounds homogeneous
additions but does not threaten the result, since external pretraining
still lifts the ensemble (\secref{performance_results}).

\noindent\textbf{No human upper bound.} No human or inter-rater ceiling
exists for body-only 12-emotion acted recognition; the official STGCN++
result is the de-facto reference, against which the
protocol-matched $+11.07$~pp lift remains the testable claim.

\noindent\textbf{Country strata.} JP/TW differences are not interpretable
as cultural effects; we report them as a stratum entangled with
capture era and performer idiolect. The five OptiTrack-captured performers
are absent from training (\secref{task_data_eval}), and excluding
the seven earliest-captured Japanese training performers moves the
ensemble Macro-F1 by only $-0.44$~pp ($36.50$ vs $36.94\%$ pooled OOF),
bounding the capture-era effect.

\noindent\textbf{Triangulation and the held dataset.} The saliency--counterfactual
agreement is $n=6$, $p=0.33$; the load-bearing cross-method result is the
$\rho=+0.500$ vs $+0.033$ contrast over the $48$ emotion--region pairs,
interval-estimated with emotion-block resampling ($p<0.001$;
\secref{explainability_results}), and it reproduces on the submitted
11-way ensemble ($\rho=+0.517$, $p=0.001$).
We built an anonymized motion$\rightarrow$rationale dataset but withhold
release pending performer-consent/licen\-se review and because skeleton
renders retain residual re-identifiability; we release the narrator method
and audit so the results are reproducible.

\section{Conclusion}
\label{sec:conclusion}

Under performer-held-out evaluation of body-only 12-class acted-emotion
recognition, reliable gains came not from a new architecture but from
combining models with orthogonal error modes, a protocol-matched
$+11.07$~pp over the same-split reproduced baseline. The harder
contribution: part-masking, stability, and counterfactual tests showed
that a strong member's decisions, and those of the submitted ensemble,
\emph{depend} on motion-grounded body-region evidence, and that this
saliency \emph{aligns} with rule-based Laban attributes, while the suite
faithfully reported a negative and a $0/50$-audited narrator. The gain is
cross-validated on the labeled training performers, and the hidden-test
score of $37.23\%$ Macro-F1 earned the Best Performance Award. Generalisation under
performer shift needs both complementary predictions and faithful
explanations of the motion evidence behind them.

\clearpage
\section*{Ethical Impact Statement}
\label{sec:ethics}

\noindent\textbf{Consent and approval.}
The DIEM-A corpus was collected by the dataset holder under their own
ethics approval and explicit performer consent, as described in the
dataset publication~\cite{cheng2025diema}. This work uses only the data released to MMAC
challenge participants; we did not collect new recordings, identifiers, or
auxiliary biometric attributes, and we did not contact or re-identify any
performer.

\noindent\textbf{Bias and limited generalizability.}
DIEM-A includes 40 Japanese and 34 Taiwanese performers in our LPO split,
so the reported findings are bounded by an East-Asian acted-affect
distribution; intercultural generalization beyond these two populations is
not tested and should not be assumed. Acted emotion further differs from
spontaneous expression along intensity, duration, and self-monitoring
axes, which leaves an acted--spontaneous gap that limits transfer to in-the-wild
affect inference. We inherit whatever demographic gaps (gender, age)
exist in the source corpus. The JP--TW Macro-F1 difference reported in
\secref{performance_results} is a confounded dataset stratum, entangled
with capture era and performer idiolect, and is not interpreted as a
cultural effect; the dataset's five OptiTrack-captured performers are
absent from our training split (\secref{task_data_eval}).

\noindent\textbf{Potential misuse and mitigation.}
Body-only affect classifiers could be misused for affect surveillance or
unconsented employment screening. We therefore release method and audit
artifacts only, not a deployable affect detector, and the corpus is
licensed for research use. As an additional disclosure to downstream
users, the same skeleton features that yield $36.80\%$ emotion Macro-F1
also support roughly $69.8\%$ performer-country classification from the
same inputs; that is, country identity is substantially more decodable
than the target emotion. This exposes a re-identification and over-fitting
risk for any deployment built on similar body-only features and motivates
the held-data decision below.

\noindent\textbf{Generalizability limits.}
The performer ceiling is small ($n=92$ across the DIEM-A challenge subset,
of which $n=74$ are used in our LPO split). Inputs are skeleton-only (no face,
audio, or scene cues), and the $64$-frame analysis window covers about
$7.5\%$ of the median sequence. Macro-F1 of $36.80\%$ on a 12-way task
(chance $8.3\%$) remains far below any plausible threshold for human-usable
affect inference. The ensemble is reported as a research probe of when
ensemble diversity helps for body-only acted affect, not as a deployable
emotion recognizer.

\noindent\textbf{Held rationale dataset and residual re-identifiability.}
The anonymized motion-to-rationale natural-language dataset prepared
alongside the explainability suite is deliberately withheld pending
performer-consent and license review with the data holder. Skeleton
renderings are derived motion: the data holder can in principle re-link a
rendered clip back to its source identity, so full unlinkability is not
achievable from artifacts in our pipeline. We therefore release only the
narrator method and the audit needed to reproduce the explainability
findings; the rationale dataset itself is held until the consent/license
review is complete.

\bibliographystyle{IEEEtran}
\bibliography{references}

\clearpage
\section*{Supplementary Material}
\setcounter{figure}{0}
\setcounter{table}{0}
\renewcommand{\thefigure}{S\arabic{figure}}
\renewcommand{\thetable}{S\arabic{table}}
\renewcommand{\theHfigure}{S\arabic{figure}}
\renewcommand{\theHtable}{S\arabic{table}}

\section*{S1. Lift Path}

\begin{figure}[t]
\centering
\includegraphics[width=\columnwidth]{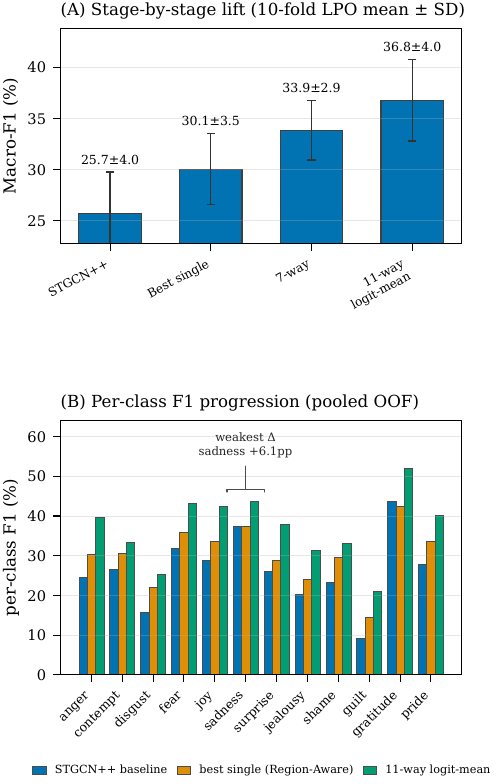}
\caption{(A) Stage-by-stage Macro-F1 over the 11-model lift path (bars = 10-fold LPO mean $\pm$ SD; sample-level paired bootstrap $95\%$ CIs, $B{=}1000$, seed~42), every stage recomputed under the canonical logit-mean convention. (B) Per-class F1 progression (pooled OOF): STGCN++ $\rightarrow$ best single (Region-Aware) $\rightarrow$ 11-way logit-mean; all 12 classes improve, weakest sadness ($+6.1$~pp).}
\label{fig:sup_lift_path}
\end{figure}

\section*{S2. Confusion and Country Strata}

\begin{figure*}[t]
\centering
\includegraphics[width=\textwidth]{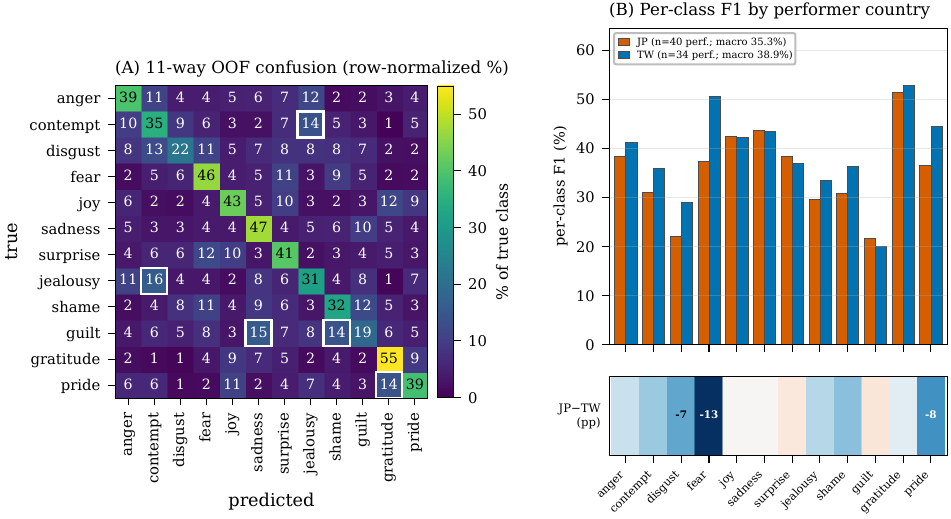}
\caption{(A) $12\times12$ row-normalised pooled-OOF confusion of the 11-way logit-mean ensemble ($n=7{,}992$ clips; rows = true class, rows sum to $100\%$); the five strongest off-diagonal confusions are boxed. (B) 11-way per-class F1 by performer country (JP $n{=}40$, TW $n{=}34$ training performers) with the signed JP$-$TW per-class strip in percentage points. The JP--TW macro gap ($35.3$ vs $38.9\%$) is a dataset stratum: the five OptiTrack-captured performers are absent from the training split, and excluding the seven earliest-captured Japanese training performers shifts pooled-OOF Macro-F1 by only $-0.44$~pp.}
\label{fig:sup_confusion_jptw}
\end{figure*}

\section*{S3. Leave-One-Out Ablations}

\begin{table}[t]
\centering
\scriptsize
\setlength{\tabcolsep}{3pt}
\caption{Leave-one-out ablation of the 11-member ensemble: change in pooled-OOF Macro-F1 (pp, $n\,{=}\,7{,}992$, 74-performer split) when one member is removed from the logit-mean fusion (last row: the whole frozen external block). Pooled-OOF base $36.94\%$ (per-fold headline: $36.80\,{\pm}\,4.00\%$). All deltas are negative.}
\label{tab:member_loo}
\begin{tabular}{@{}l l r@{}}
\toprule
Member & Family & $\Delta$ Macro-F1 (pp) \\
\midrule
MAMP-NTU60 & External (frozen) & $-0.84$ \\
SkateFormer & Attention & $-0.68$ \\
MAMP-NTU120 & External (frozen) & $-0.61$ \\
Region-Aware ConvTr & Graph-conv. & $-0.59$ \\
C3D-marker-stats & External (frozen) & $-0.52$ \\
MotionBERT-Lite & External (frozen) & $-0.47$ \\
Keypoint-Pool-MLP & Hybrid/MLP & $-0.45$ \\
STGCN++ & Graph-conv. & $-0.42$ \\
Conv1D+Transformer & Hybrid/MLP & $-0.30$ \\
ProtoGCN & Graph-conv. & $-0.15$ \\
CTR-GCN & Graph-conv. & $-0.09$ \\
\midrule
All frozen externals (4) & External (frozen) & $-2.91$ \\
\bottomrule
\end{tabular}
\end{table}

\begin{table*}[t]
\centering
\scriptsize
\setlength{\tabcolsep}{4pt}
\caption{Per-stratum leave-one-out deltas (pp, pooled OOF) for the 11-member logit-mean ensemble. `Excl.\ earliest JP' scores after removing the seven earliest-captured Japanese training performers (JP\_06--JP\_12).}
\label{tab:sup_member_loo_strata}
\begin{tabular}{@{}l l r r r r@{}}
\toprule
Member & Family & All & JP & TW & Excl.\ earliest JP \\
\midrule
MAMP-NTU60 & External (frozen) & $-0.84$ & $-1.65$ & $-0.01$ & $-0.84$ \\
SkateFormer & Attention & $-0.68$ & $-0.48$ & $-0.93$ & $-0.78$ \\
MAMP-NTU120 & External (frozen) & $-0.61$ & $-0.90$ & $-0.30$ & $-0.62$ \\
Region-Aware ConvTr & Graph-conv. & $-0.59$ & $-0.62$ & $-0.57$ & $-0.65$ \\
C3D-marker-stats & External (frozen) & $-0.52$ & $-0.64$ & $-0.42$ & $-0.54$ \\
MotionBERT-Lite & External (frozen) & $-0.47$ & $-0.72$ & $-0.23$ & $-0.49$ \\
Keypoint-Pool-MLP & Hybrid/MLP & $-0.45$ & $-0.62$ & $-0.28$ & $-0.44$ \\
STGCN++ & Graph-conv. & $-0.42$ & $-0.67$ & $-0.14$ & $-0.44$ \\
Conv1D+Transformer & Hybrid/MLP & $-0.30$ & $-0.40$ & $-0.17$ & $-0.27$ \\
ProtoGCN & Graph-conv. & $-0.15$ & $-0.10$ & $-0.19$ & $-0.20$ \\
CTR-GCN & Graph-conv. & $-0.09$ & $-0.15$ & $-0.05$ & $-0.05$ \\
\midrule
All frozen externals (4) & External (frozen) & $-2.91$ & $-3.11$ & $-2.75$ & $-2.74$ \\
\bottomrule
\end{tabular}
\end{table*}

\section*{S4. LMA Schema}

\begin{table*}[t]
\centering
\caption{Rule-based Laban Movement Analysis (LMA) attribute schema: 32
clip-level features across the four Laban axes, computed from BVH-24 forward
kinematics. The per-emotion z-score signature and the region aggregation
$\ell_{c,r}$ are defined in the main paper; this is
the established movement vocabulary the saliency alignment is measured
against.}
\label{tab:sup_lma_schema}
\scriptsize
\begin{tabular}{l p{4.2cm} p{9.6cm}}
\toprule
Laban axis & What it captures & Attributes (8 per axis) \\
\midrule
Body & posture and bilateral form & head bow, trunk lean, arm openness, left--right speed asymmetry, left--right position asymmetry, stillness ratio, head lateral tilt, shoulder drop \\
Effort & motion quality (time/weight/flow) & suddenness, sustainedness, strong energy, light energy, bound flow, free flow, intensity peak, intensity variance \\
Shape & body volume and its change & body volume, shoulder width, head height, contraction change, rise/sink, spread change, arm enclosure, advance/recede \\
Space & trajectory and locomotion & directness, path curvature, horizontal root displacement, vertical root displacement, lateral sway, cumulative turn, dominant direction, locomotion ratio \\
\bottomrule
\end{tabular}
\\[2pt] {\scriptsize Region importance is computed from the per-emotion z-scores defined in the main paper as the within-region mean of $|z_{c,k}|$; Spearman correlation over the 12 $\times$ 4 emotion-region pairs gives $\rho=+0.500$ vs.\ $+0.033$ for classical kinematics.}
\end{table*}

\section*{S5. Saliency}

\begin{figure*}[t]
\centering
\includegraphics[width=\textwidth]{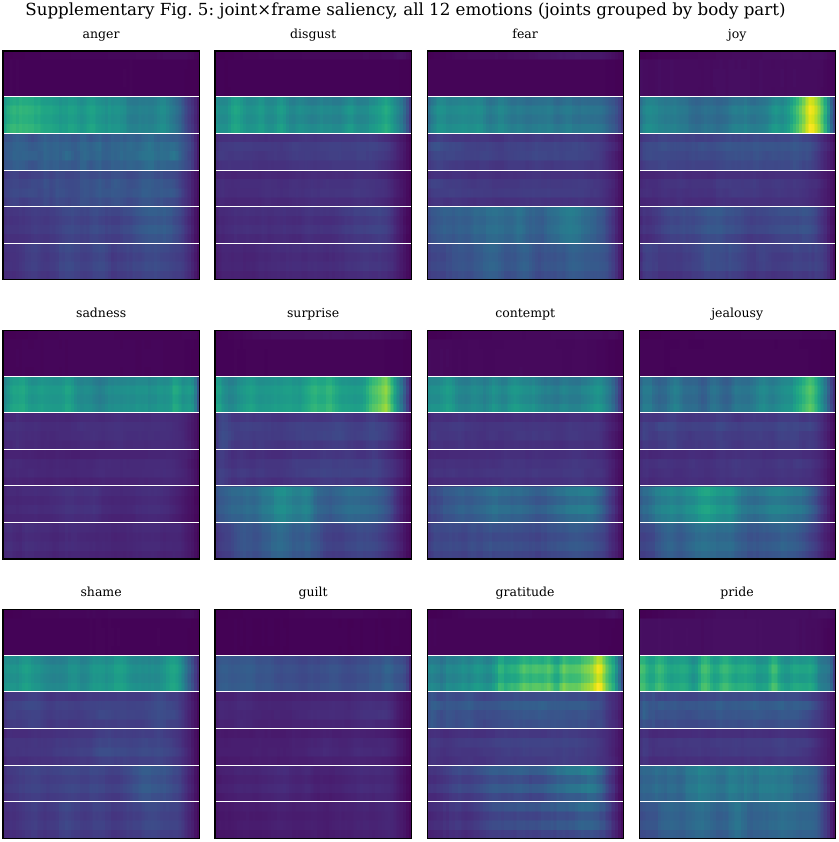}
\caption{Per-emotion temporal and spatial saliency, full grid (companion to the temporal-saliency negative reported in the main paper).}
\label{fig:sup_full_saliency_grid}
\end{figure*}

\begin{figure*}[t]
\centering
\includegraphics[width=\textwidth]{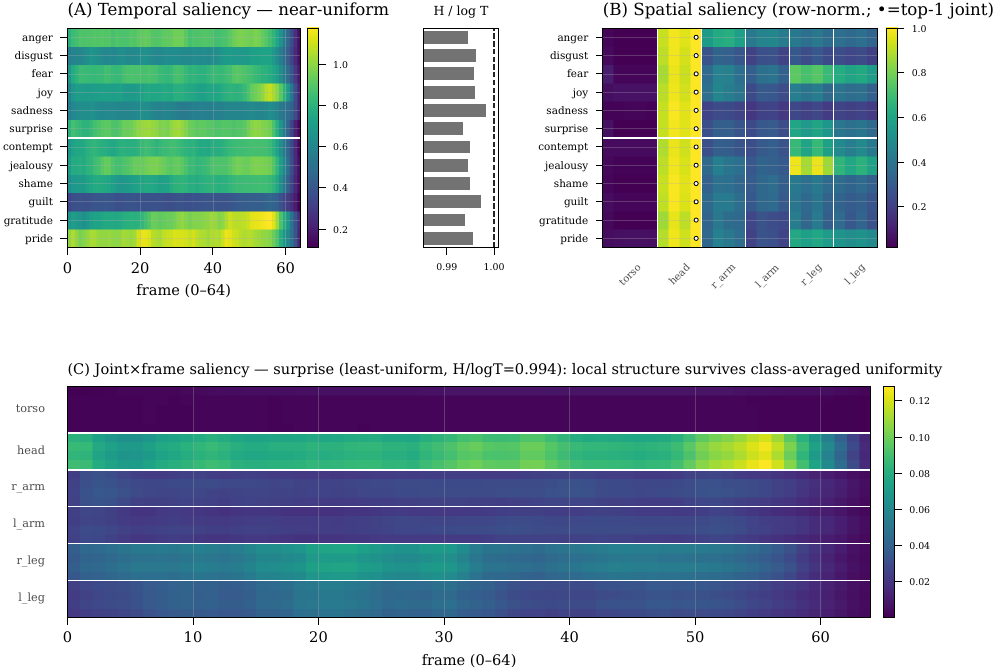}
\caption{(A) Temporal saliency (mean over joints) is near-uniform for all 12 emotions; per-sample entropy $\approx 98.7\%$ of $\log(64)$ (class-mean $99.5\%$), the load-bearing negative. (B) Spatial saliency (row-normalised, joints in 6 body parts): Head is the top-1 joint for all 12 emotions (Head/Neck/Neck1 dominate). (C) For surprise (least-uniform), a joint$\times$frame burst shows local structure survives class-averaged temporal uniformity. Colourblind-safe (viridis).}
\label{fig:sup_saliency}
\end{figure*}

\section*{S6. LMA and Counterfactuals}

\begin{figure*}[t]
\centering
\includegraphics[width=\textwidth]{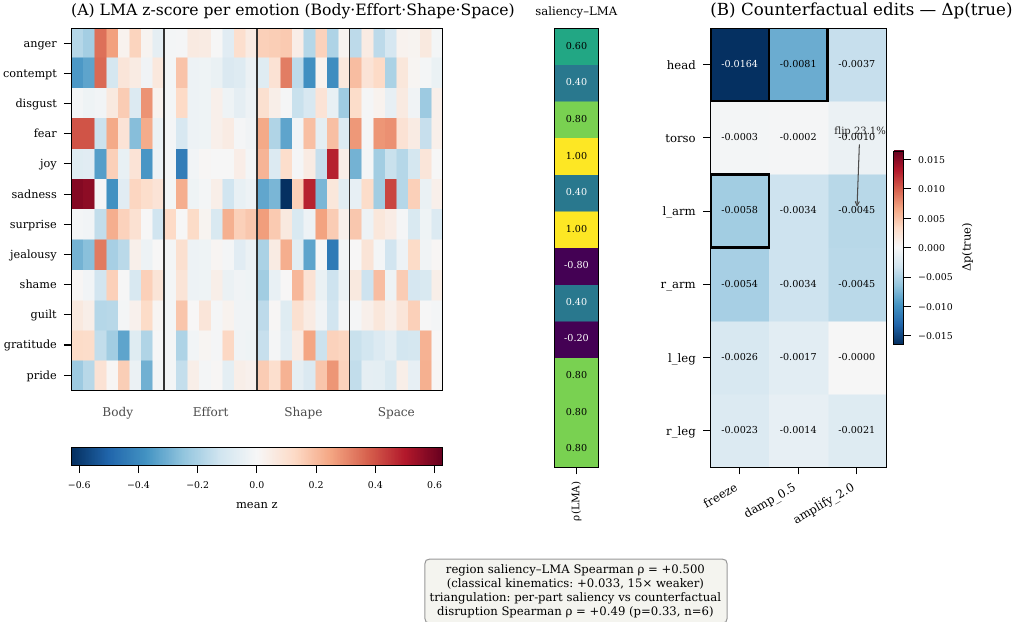}
\caption{(A) Mean LMA z-score per emotion ($32$ attrs: Body/Effort/Shape/Space); right strip = per-emotion saliency--LMA Spearman. Region saliency--LMA $\rho=+0.500$ vs $+0.033$ for classical kinematics ($15\times$ weaker). (B) Counterfactual motion edits (observational perturbations, not causal): mean $\Delta p_{\text{true}}$ over $864$ val $\times$ 10 folds; head-freeze most disruptive ($-0.0164$, $41\%$ flip), l\_arm+amplify $23.1\%$ flip. Per-part saliency vs disruption Spearman $\rho=+0.49$ ($n=6$, $p=0.33$), a positive but modest triangulation.}
\label{fig:sup_lma_counterfactual}
\end{figure*}

\section*{S7. Qualitative Cards}

\begin{figure}[t]
\centering
\includegraphics[width=\columnwidth]{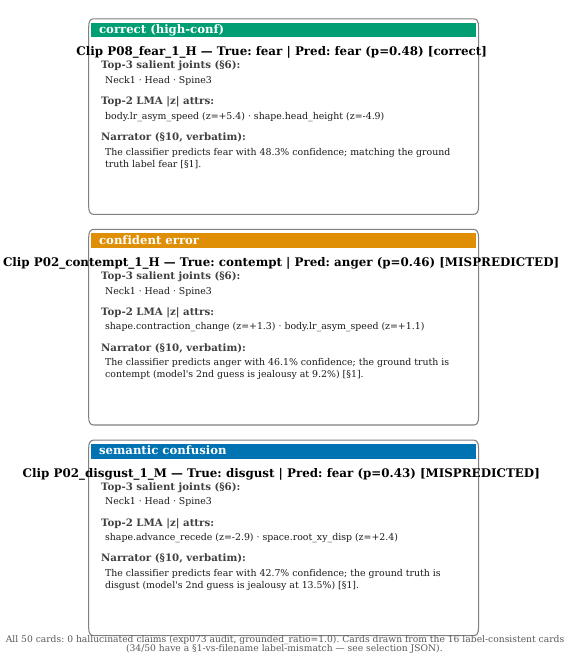}
\caption{Three explanation cards (correct / confident-error / semantic-confusion) from the $16$ label-consistent cards; each cites source channels (salient joints, top-2 LMA $|z|$, verbatim deterministic narrator). All 50 cards pass the grounding audit ($0$ hallucinated claims, grounded-ratio $=1.0$). Qualitative; per-card correctness is excluded pending a card-generator fix.}
\label{fig:sup_cards}
\end{figure}

\section*{S8. Full Negative-Results Catalogue}

\begin{table*}[t]
\centering
\caption{Supplementary: full catalogue of negative results (extends Table 2).}
\label{tab:sup_table2_negatives_full}
\footnotesize
\begin{tabular}{l p{2.6cm} p{2.7cm} p{4.9cm}}
\toprule
Approach & Configuration & Outcome ($\Delta$F1) & Methodological lesson \\
\midrule
Supervised contrastive head & pairwise SupCon, $\lambda$=0.05, batch 128 & -11.53 pp, significant & 12 classes at batch 128 give too few positives per class, suggesting a per-class center loss may fit better at this scale \\
Mixture-of-experts fusion & K=4, hand-crafted gradient-free routing & collapses to 6.55\% F1 & a shared head with random init and 21-D routing is associated with backbone collapse, suggesting a feature adapter, learned routing, and zero-init are needed \\
Scenario-text alignment & global InfoNCE, 3-fold & -1.42 pp & the projection collapses to performer identity (cross-group retrieval 10.6\%) \\
Part-rationale alignment & cosine text alignment, 3-fold & -0.64 pp, CI [-1.30, -0.09] & the text bottleneck is associated with an F1 ceiling of 9--12\%; shortcut-breaking works but yields no F1 lift \\
VLM zero-shot distillation & 12-class, a recent VLM, preliminary & failed in preliminary zero-shot ranking & the true class ranked last (12/12); a vision-language model did not infer emotion reliably from faceless, scene-free skeleton renderings \\
Contact-marker late fusion & C3D 22-D contact stats, 3-fold & -0.59 pp & the contact features carry a 69.8\% country signal, so some performers go out of distribution under LPO, suggesting a domain-invariance objective is needed \\
Multi-crop window inference & 24-setting window sweep & -12 to -17 pp & the model is trained on full-span subsamples, so compact windows are a distribution shift \\
Post-hoc calibration & nested-LPO, 7-way ensemble & 0 to -0.14 pp, not significant & equal-weight argmax is already near-optimal; bias-style calibration did not exceed it in this search \\
Same-architecture stacking & 8-way (six models + two ConvTr variants) & -0.29 pp & diversity saturates when stacking variants of a single architecture \\
Over-stacked SSL ensemble & 13-way, three-plus MAMP checkpoints & -0.88 pp & one SSL family over-stacked dilutes the production members; two MAMP checkpoints is the sweet spot \\
Arithmetic-mean averaging & probability-domain averaging & -1.15 pp (forgone) & the canonical convention is logit-mean (raw-logit averaging); the arithmetic alternative leaves -1.15 pp on the table \\
PoseC3D heatmap 3D-CNN & preliminary evaluation & 20.26\% (below the 24\% threshold) & architectural novelty is not error-space orthogonality; measure the contribution in error space \\
Joint-masking sparse model & options A/B, 3-fold & -0.33 pp (B); collapse (A) & attribution-driven sparse models are not ensemble-additive (the 7-way already sees those joints) \\
Joint-subset training & top-N joints only, 3-fold & top-4 19.83\%, CI [17.67, 21.99] & input-driven sparsity converges to the same negative as joint-masking \\
Heavy regularisation, short horizon & Conv1D+Transformer, 40 epochs & -1.34 pp standalone, CI [-0.32, +0.35] & heavy regularisation needs a long horizon; it under-fits at 40 epochs \\
Focal loss & single-model variant & -0.48 F1 / -0.99 Acc & the class-balanced DIEM-A does not benefit from focal loss \\
SkateFormer with SGD (lr=0.2) & single-model variant & chance-level collapse & transformer backbones require AdamW \\
Joint dropout (0.1) & single-model variant & seed-dependent fold collapse & some folds collapse under some seeds; not adoptable \\
Long clip length (128) & STGCN++, clip 128 vs 64 & regresses & the emotion signal concentrates in a short temporal window \\
Content/style dual head & GRL + style loss, default weights & -2.65 pp & GRL and a style loss obstruct backbone learning at default weights \\
Gradient temporal saliency & explainability probe & AUC gap +0.003 ($\approx$0), not significant & within-window frame importance is near-uniform (entropy: per-sample 98.7\%, class-mean 99.5\% of log T); use the body-part axis instead \\
\bottomrule
\end{tabular}
\\[2pt] \parbox{\linewidth}{\raggedright\scriptsize Full negative-results catalogue (supplementary); the main paper shows the curated eight (Table 2).}
\end{table*}


\section*{S9. Member Error Correlation}

\begin{figure}[t]
\centering
\includegraphics[width=\columnwidth]{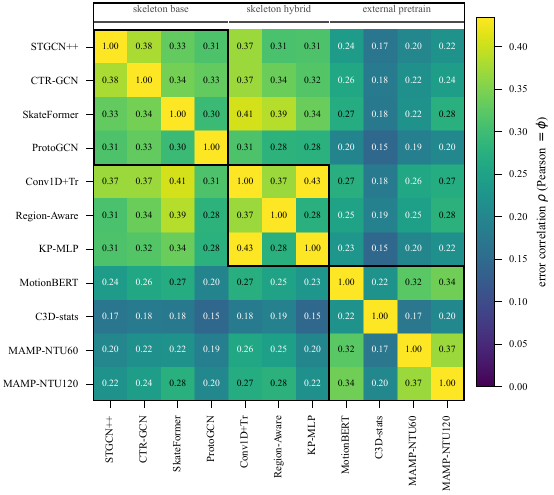}
\caption{Members make largely independent errors, the diversity the ensemble converts into its gain: off-diagonal error correlations stay in $[0.15, 0.43]$, lowest for the external-pretrain C3D-stats column ($\rho \approx 0.15$--$0.19$ to skeleton models); most-redundant pair Conv1D+Tr\,$\leftrightarrow$\,KP-MLP ($\rho=0.43$). Pairwise Pearson ($=\phi$), OOF $n=7{,}992$; the diagonal is 1 by definition and the colour scale is capped at the off-diagonal maximum. KP-MLP\,$=$\,Keypoint-Pool-MLP; C3D-stats\,$=$\,C3D marker statistics, not a 3D CNN.}
\label{fig:sup_member_corr}
\end{figure}

\end{document}